\documentclass[10pt,twocolumn,letterpaper]{article}

\usepackage{cvpr}
\usepackage{times}
\usepackage{epsfig}
\usepackage{graphicx}
\usepackage{amsmath}
\usepackage{amssymb}
\usepackage{booktabs}
\usepackage{setspace}
\usepackage{enumitem}

\usepackage[breaklinks=true,bookmarks=false]{hyperref}

\cvprfinalcopy 

\def\cvprPaperID{1225} 

\begin{document}

\title{Cross-domain Detection via Graph-induced Prototype Alignment}

\author{
	Minghao Xu\textsuperscript{\rm 1,2} \quad
	Hang Wang\textsuperscript{\rm 1,2} \quad
	Bingbing Ni\textsuperscript{\rm 1,2,3}\footnotemark[1] \quad
	Qi Tian\textsuperscript{\rm 4} \quad
	Wenjun Zhang\textsuperscript{\rm 1}\\
	\textsuperscript{\rm 1}Shanghai Jiao Tong University, Shanghai 200240, China \\
	\textsuperscript{\rm 2}MoE Key Lab of Artificial Intelligence, AI Institute, Shanghai Jiao Tong University \\
	\textsuperscript{\rm 3}Huawei Hisilicon \quad
	\textsuperscript{\rm 4}Huawei Noah’s Ark Lab\\
	\{xuminghao118, wang--hang, nibingbing, zhangwenjun\}@sjtu.edu.cn \\
	nibingbing@hisilicon.com \quad
	tian.qi1@huawei.com
}

\maketitle
\thispagestyle{empty}


\begin{abstract}
	\vspace{-2mm}
	Applying the knowledge of an object detector trained on a specific domain directly onto a new domain is risky, as the gap between two domains can severely degrade model's performance. Furthermore, since different instances commonly embody distinct modal information in object detection scenario, the feature alignment of source and target domain is hard to be realized. To mitigate these problems, we propose a Graph-induced Prototype Alignment (GPA) framework to seek for category-level domain alignment via elaborate prototype representations. In the nutshell, more precise instance-level features are obtained through graph-based information propagation among region proposals, and, on such basis, the prototype representation of each class is derived for category-level domain alignment. In addition, in order to alleviate the negative effect of class-imbalance on domain adaptation, we design a Class-reweighted Contrastive Loss to harmonize the adaptation training process. Combining with Faster R-CNN, the proposed framework conducts feature alignment in a two-stage manner. Comprehensive results on various cross-domain detection tasks demonstrate that our approach outperforms existing methods with a remarkable margin. Our code is available at \url{https://github.com/ChrisAllenMing/GPA-detection}.
	\vspace{-3mm}
\end{abstract}


\section{Introduction} \label{sec1}

Following the rapid development of techniques leveraging Deep Neural Networks (DNNs)\footnotetext[0]{*The corresponding author is Bingbing Ni.}, a variety of computer-vision-related tasks, \emph{e.g.} object classification \cite{alexnet,resnet}, object detection \cite{faster_rcnn,ssd}, and semantic segmentation \cite{deeplab,mask_rcnn}, witnessed major breakthroughs in the last decade. It should be noticed that the impressive performance of these models is established, to a great extent, on the basis of massive amounts of annotated data, of which the annotation process itself could be a laborious task in many cases. Furthermore, when the model trained on a domain with abundant annotations is applied to a distinct domain with limited, even unavailable, labels, it will suffer from performance decay, due to the existence of domain shift \cite{deep_transferrable}. 


\begin{figure}[t]
	\centering
	\includegraphics[width=0.47\textwidth]{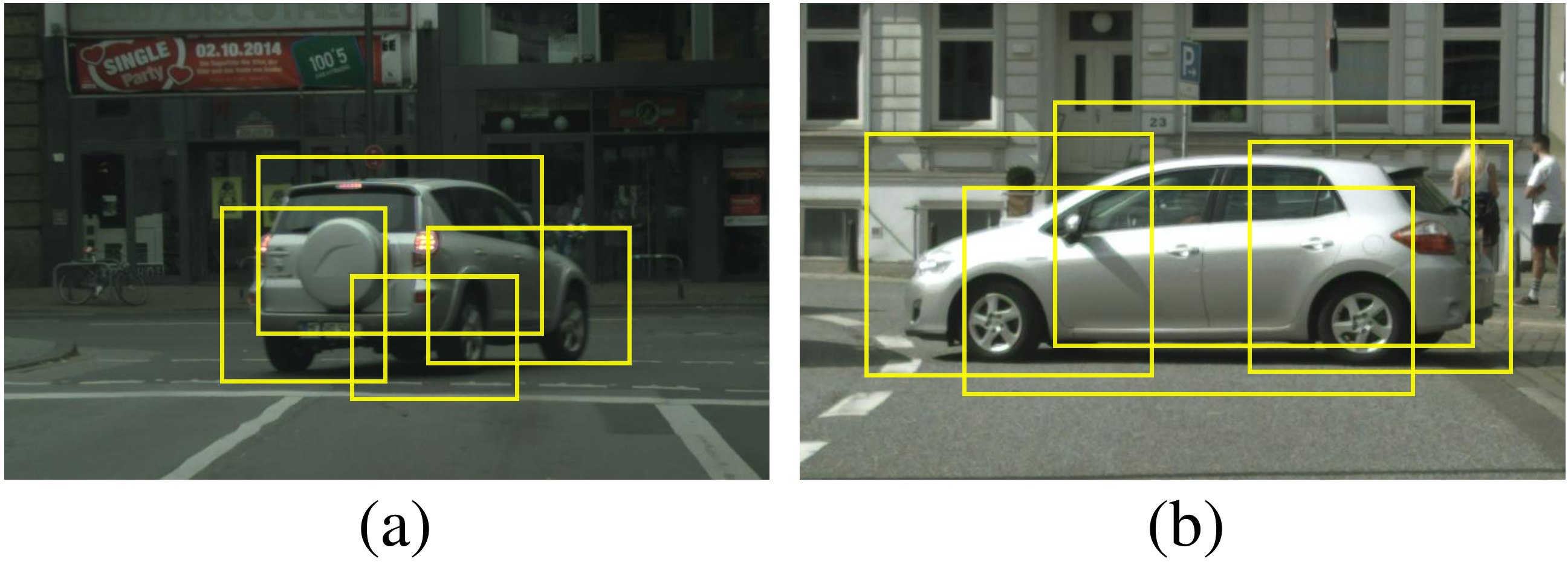}
	\caption{Two vehicles and corresponding region proposals from the Cityscapes \cite{city} dataset which serves as target domain. These two vehicles reflect multi-modal information, \emph{e.g.} distinct scale and orientation, and the generated region proposals contain incomplete information of them.} 
	\label{fig_motivation}
	\vspace{-5mm}
\end{figure}


One of the extensively explored techniques to deal with such dilemma is Unsupervised Domain Adaptation (UDA), which seeks for knowledge transfer from a labeled dataset (source domain) to another unlabeled one (target domain). In order to encourage domain-invariant feature representations, commonly adopted strategies can be roughly classified into two categories: 1) Minimizing an explicitly defined domain discrepancy measurement \cite{dan,domain_confusion,deepcoral,wdan}; 2) Applying adversarial training to UDA via domain classifier \cite{revgrad,adda,pixel-level,multi_adversarial}. These strategies are comprehensively exploited in classification-based tasks. 

Besides classification, cross-domain detection is also strongly demanded in modern Computer Vision systems, including intelligent surveillance and autonomous driving, in which the deployment environment, \emph{e.g.} backgrounds, weather, illumination, changes from site to site. Previous works \cite{da_faster_rcnn,mean_teacher,adapting_detector} utilize independent or grouped region proposals to align source and target domain on local instance level. However, since supervisory signal is lacked on target domain, the generated region proposals commonly deviate from instances, which makes the information from primal proposals improper to depict corresponding instances. In addition, the representation of an instance is insufficient to characterize the category it belongs to, because a single instance can only reflect limited modal information, \emph{e.g.} specific scale or orientation. However, the representations of instances within a category are multi-modal. Two typical examples are illustrated in Figure \ref{fig_motivation}, where two vehicles express different modal information, and the generated region proposals deviate from objects. These two problems make instance-level domain alignment trapped into dilemma. Except for these issues, in multi-class cross-domain detection tasks, class-imbalance leads to the inconsistency of domain adaptation process among different classes along training, which greatly impairs model's adaptation performance on those sample-scarce categories.

Motivated by these problems, we propose the \textbf{Graph-induced Prototype Alignment} (GPA) framework and embed it into a two-stage detector, Faster R-CNN \cite{faster_rcnn}. For the sake of better local alignment via region proposals, we introduce two key components, \emph{graph-based region aggregation} and \emph{confidence-guided merging}. In graph-based region aggregation, a relation graph which takes both the location and size of proposals into consideration is constructed to aggregate features on instance level, such that the critical features of each instance are integrated. In confidence-guided merging, the multi-modal information contained in various instances is embodied by  prototype\footnotemark[2]\footnotetext[2]{Prototype is the representative embedding of all samples within the same class.} representations, such that, by utilizing the complementarity of multi-modal information, each category can be better characterized. Using prototypes as the proxy of different classes, category-level domain alignment is performed. Furthermore, considering that class-imbalance exists in the multi-class cross-domain detection tasks, we harmonize the process of domain adaptation via a \emph{Class-reweighted Contrastive Loss}, in which the sample-scarce classes are assigned with higher weights, thus they can be better aligned during training. 

Based on the two-stage structure of Faster R-CNN, we also conduct feature alignment in a two-stage manner: 1) In the first stage, foreground and background distributions are separated, and class-agnostic alignment is performed on feature distributions of two domains; 2) In the second stage, more fine-grained alignment is respectively performed on each foreground category.

Our contributions can be summarized as follows:
\vspace{-1mm}
\begin{itemize}
	\item We propose the Graph-induced Prototype Alignment (GPA) framework, in which more precise instance-level features are obtained through graph-based region aggregation, and prototype representations are derived for category-level domain alignment.
	\vspace{-1mm}
	\item In multi-class cross-domain detection tasks, for tackling the class-imbalance during feature alignment, we design a Class-reweighted Contrastive Loss to harmonize the adaptation process among different classes.
	\vspace{-2mm}
	\item Combining with the Faster R-CNN architecture, we propose a two-stage domain alignment scheme, and it achieves state-of-the-art performance on the cross-domain detection tasks under various scenarios.
\end{itemize}


\section{Related Work} \label{sec2}

\par{\textbf{Object Detection.}}
Current object detection methods can be roughly categorized into two classes: one-stage detectors \cite{yolo,ssd,yolo9000,focal_loss} and two-stage detectors \cite{rcnn, fast_rcnn, faster_rcnn, FPN,mask_rcnn}. R-CNN \cite{rcnn} first obtains region proposals with selective search and then classifies each proposal. Fast R-CNN \cite{fast_rcnn} speeds up detection process by introducing RoI pooling. Faster R-CNN \cite{faster_rcnn} produces nearly cost-free region proposals with Region Proposal Network. One-stage detectors, such as YOLO \cite{yolo} and SSD \cite{ssd}, directly predict category confidence and regress bounding box based on predefined anchors. Lin \emph{et al.} \cite{focal_loss} proposed focal loss to address class-imbalance, which increases the accuracy of one-stage detector. In this work, we choose Faster R-CNN as baseline detector for its robustness and scalability.


\begin{figure*}[t]
	\centering
	\includegraphics[width=1.0\textwidth]{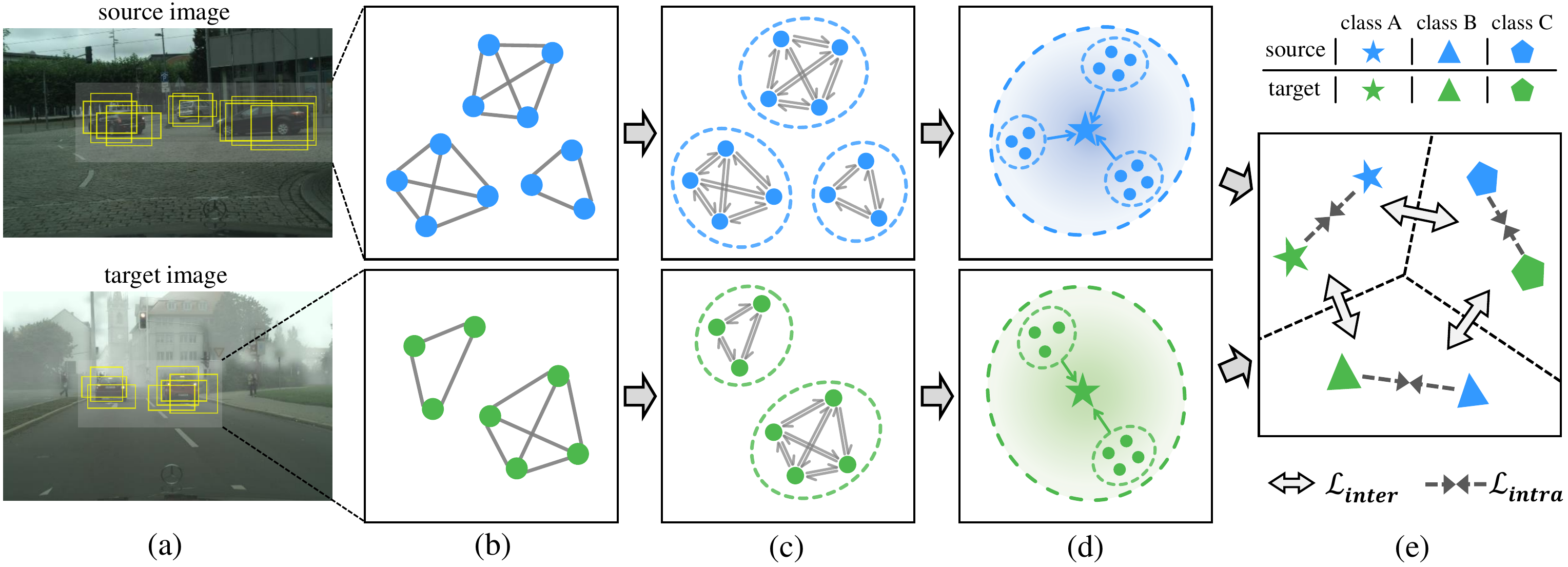}
	\caption{\textbf{Framework overview.} (a) Region proposals are generated. (b) Constructing the relation graph on produced region proposals. (c) More accurate instance-level feature representations are obtained through information propagation among proposals belonging to the same instance. (d) Prototype representation of each class is derived via confidence-guided merging. (e) Performing category-level domain alignment through enhancing intra-class compactness and inter-class separability.} 
	\label{fig_framework}
	\vspace{-4mm}
\end{figure*}


\par{\textbf{Unsupervised Domain Adaptation (UDA).}}
UDA aims to generalize the model learned from labeled source domain to the other unlabeled target domain. In the field of UDA, a group of approaches focus on minimizing a specific domain discrepancy metric, \emph{e.g.}, Maximum Mean Discrepancy (MMD) \cite{two-sample_test,domain_confusion}, Weighted MMD \cite{w-mmd}, Multi-Kernel MMD \cite{dan} and Wasserstein Distance \cite{w-distance}. Another research line is based on adversarial training, in which a domain classifier is introduced to facilitate domain-invariance on feature level \cite{revgrad,adda,cdan} or pixel level \cite{gta,cycada,domain_mixup}. Recently, several works \cite{semantic, collaborative, transferrable_proto, progressive} utilize pseudo labels of samples from target domain to introduce discriminative information during domain alignment. Following the prototype-based approaches \cite{semantic,transferrable_proto}, we extend the usage of prototype to cross-domain detection tasks.


\par{\textbf{Cross-domain Detection.}}
Beginning with the work of Chen \emph{et al.} \cite{da_faster_rcnn}, the topic of cross-domain detection arouses interests in the community of UDA. In that work, a Domain Adaptive Faster R-CNN model is constructed to reduce domain discrepancy on both image and instance levels. More recently, Saito \emph{et al.} \cite{strong-weak} proposed a strong-weak alignment strategy which puts less effort on aligning globally dissimilar images. Cai \emph{et al.} \cite{mean_teacher} remolded the mean teacher scheme for cross-domain detection. Kim \emph{et al.} \cite{diversify_and_match} used domain diversification to learn feature representations which are invariant among multiple domains. Zhu \emph{et al.} \cite{adapting_detector} solved the questions of ``where to look" and ``how to align" via two key components, region mining and region-level alignment. In \cite{robust_learning}, domain adaptation problem is tackled from the perspective of robust learning. 

\emph{Improvements over existing methods.} Although former works \cite{da_faster_rcnn,mean_teacher,adapting_detector} seek for instance-level domain alignment using region proposals, they fail to derive exact instance-level representations and ignore the multi-modal information of various instances. In this work, we utilize relation graph to obtain more precise instance-level feature representations, and per-category prototypes are derived to integrate different instances' multi-modal information. 


\par{\textbf{Graph Convolutional Network (GCN).}}
GCN \cite{gcn_model} has been explored as a manner to learn graph relations with convolution, which boosts the optimization of graph-based model. Because of the effectiveness and interpretability of GCN, it has been widely applied to various tasks, \emph{e.g.}, action recognition \cite{graph_action_recognition}, person Re-ID \cite{graph_reid_yan}, video understanding \cite{graph_video,graph_video_caption} and point cloud learning \cite{graph_point_cloud}. Several recent works \cite{adagraph,gcan} utilize graph model to structure multiple domains and categories for classification-based domain adaptation. For cross-domain detection, we employ graph structure to model the relation among region proposals.


\section{Method} \label{sec3}

In Unsupervised Domain Adaptation (UDA), source domain $\mathcal{S} = \{(x^\mathcal{S}_i, y^\mathcal{S}_i)\}^{N_\mathcal{S}}_{i=1}$ is characterized by $N_\mathcal{S}$ i.i.d. labeled samples, where $x^\mathcal{S}_i$ follows source distribution $\mathbb{P}_\mathcal{S}$ and $y^\mathcal{S}_i$ denotes its corresponding label. Similarly, target domain $\mathcal{T} = \{x^\mathcal{T}_j\}^{N_\mathcal{T}}_{j=1}$ is represented by $N_\mathcal{T}$ i.i.d. unlabeled samples, where $x^\mathcal{T}_j$ follows target distribution $\mathbb{P}_\mathcal{T}$. 


\subsection{Motivation and Overview} \label{sec3_1}
\vspace{-2mm}
In contrast to domain adaptation in classification, its application in object detection is more sophisticated. In specific, since supervisory signal is lacked on target domain, foreground instances are normally represented by a bunch of inaccurate region proposals. In addition, different instances in various scenes commonly reflect diverse modal information, which makes it harder to align source and target domain on local instance level.
Another problem impairing model's performance on cross-domain detection tasks is class-imbalance. Concretely, those categories with abundant samples are trained more sufficiently, thus better aligned, while the sample-scarce categories can't be readily aligned for the lack of adaptation training.

To address above issues, we propose the \emph{Graph-induced Prototype Alignment} (GPA) framework.
In specific, domain adaptation is realized via aligning two domains' prototypes, in which the critical information of each instance is aggregated via graph-based message propagation, and the multi-modal information reflected by different instances is integrated into per-category prototypes. On the basis of this framework, \emph{Class-imbalance-aware Adaptation Training} is proposed to harmonize the domain adaptation process among different classes through assigning higher weights to the sample-scarce categories.


\subsection{Graph-induced Prototype Alignment} \label{sec3_2}

In the proposed framework, five steps are performed to align source and target domain with category-level prototype representations, just as shown in Figure \ref{fig_framework}.


\par{\textbf{Region proposal generation.}} In Faster R-CNN \cite{faster_rcnn}, region proposals are generated by Region Proposal Network (RPN) to characterize foreground and background.
These proposals provide abundant information of various instance patterns and scene styles, while they usually contain incomplete information of instances because of the deviation of bounding boxes, especially on target domain. Subsequent operations aim to extract the exact information of each instance from region proposals. 


\begin{figure}[t]
	\centering
	\includegraphics[width=0.45\textwidth]{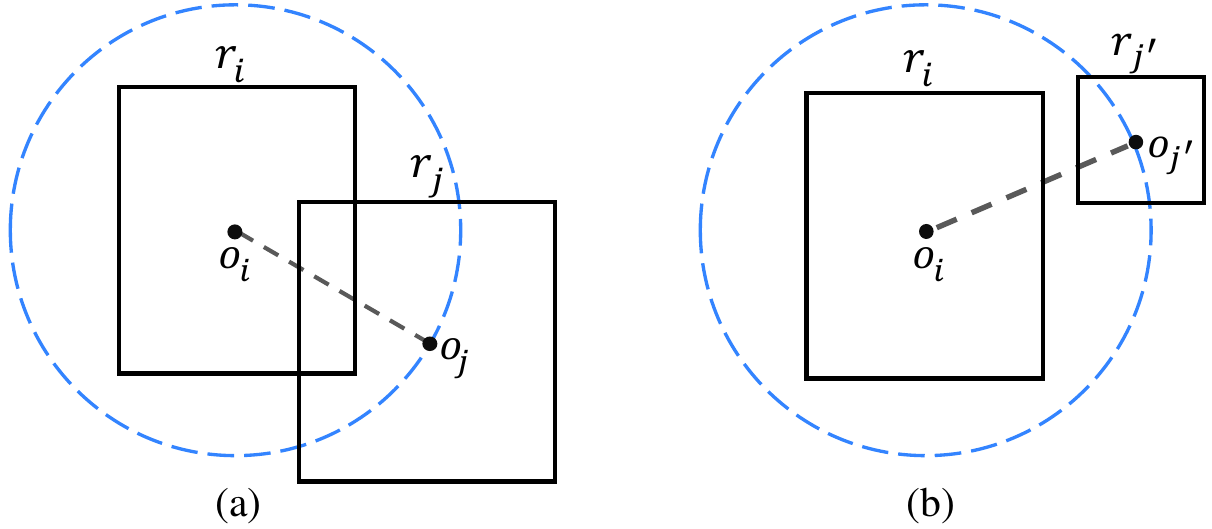}
	\caption{ Region proposal $r_i$ interacts with another two region proposals, $r_j$ and $r_{j'}$, with different sizes. } 
	\label{fig_proposal}
	\vspace{-3mm}
\end{figure}


\par{\textbf{Constructing relation graph.}} We structure the proposals generated by RPN as a graph $\mathcal{G} = (\mathcal{V}, \mathcal{E})$, where $\mathcal{V}$ represents the set of vertices corresponding to $N_p$ proposals, and $\mathcal{E} \subseteq \mathcal{V} \times \mathcal{V}$ denotes the set of edges, \emph{i.e.} the relations between proposals. Adjacency matrix $\textbf{A} \in \mathbb{R}^{N_p \times N_p}$ is used to model such relationship. Intuitively, two spatially closer proposals more likely depict the same object and should be assigned with higher connection weight. Following this intuition, a manner to obtain adjacency matrix is to apply a Gaussian kernel over the Euclidean distance between the centers of two proposals:
\vspace{-1mm}
\begin{equation} \label{eq1}
\textbf{A}_{i,j} = \textrm{exp} \Big ( - \frac{|| o_i - o_j ||^2_2}{2 \sigma^2} \Big ) ,  
\end{equation}
where $o_i$ and $o_j$ denote the centers of the $i$-th and $j$-th proposal ($1 \leqslant i, j \leqslant N_p$), and $\sigma$ is the standard deviation parameter which controls the sparsity of $\textbf{A}$. 

However, when calculating the adjacency matrix, it is unreasonable to treat proposals with various spatial sizes equally. Just as shown in Figure \ref{fig_proposal}, though region proposal pairs $(r_i, r_j)$ and $(r_i, r_{j'})$ have the equal center distance, their strength of relevance is obviously distinct, and $(r_i, r_j)$ should possess higher connection weight in $\textbf{A}$ for the larger overlap between $r_i$ and $r_j$. Intersection over Union (IoU) is a broadly used metric which takes both the location and size of proposals into consideration, and the derivation of adjacency matrix with IoU is as follows:
\begin{equation} \label{eq2}
\textbf{A}_{i, j} = \textrm{IoU} (r_i, r_j) = \frac{r_i \bigcap r_j}{r_i \bigcup r_j} , 
\end{equation}
where $r_i$ and $r_j$ denote the $i$-th and $j$-th region proposal respectively ($1 \leqslant i, j \leqslant N_p$). The setup of relation graph lays the foundation for information propagation among region proposals. The comparison between above two methods of constructing adjacency matrix is presented in Sec. \ref{sec5_1}. 


\par{\textbf{Graph-based region aggregation.}} Because of the deviation of bounding boxes, region proposals often distribute around the ground truth objects, which leads to the inaccuracy of representing an object with single proposal. In fact, primal region proposals express incomplete information of instances. In order to achieve exact instance-level feature representations, the embeddings of proposals belonging to a certain instance should be aggregated. By utilizing the spatial relevance provided by adjacency matrix $\textbf{A}$, proposals' feature embeddings $\textbf{F} \in \mathbb{R}^{N_p \times d}$ ($d$ is the dimension of embedding) and classification confidence $\textbf{P} \in \mathbb{R}^{N_p \times N_c}$ ($N_c$ is the number of classes) are aggregated as follows:
\begin{equation} \label{eq3}
\widetilde{\textbf{F}} = \textbf{D}^{-\frac{1}{2}} \textbf{A} \textbf{D}^{-\frac{1}{2}} \textbf{F} ,
\end{equation}
\begin{equation} \label{eq4}
\widetilde{\textbf{P}} = \textbf{D}^{-\frac{1}{2}} \textbf{A} \textbf{D}^{-\frac{1}{2}} \textbf{P} ,
\end{equation}
where $\textbf{D} \in \mathbb{R}^{N_p \times N_p}$ denotes the diagonal degree matrix with entries $\textbf{D}_{ii} = \sum_{j} \textbf{A}_{ij}$. In Eqs. \ref{eq3}, \ref{eq4}, after region aggregation, $\widetilde{\textbf{F}} \in \mathbb{R}^{N_p \times d}$ and $\widetilde{\textbf{P}} \in \mathbb{R}^{N_p \times N_c}$ express more precise instance-level information through information propagation among adjacent proposals.
Compared with the conventional graph convolution, we leave the learnable parameter matrix out, considering that explicit supervisory signal is lacked on the branch of domain adaptation learning.
We illustrate the benefit of such operation in Sec. \ref{sec5_1}.


\par{\textbf{Confidence-guided merging.}} Now that the feature representations are aggregated on instance level, we would like to integrate the multi-modal information reflected by different instances into prototype representations. In order to 
highlight the modal information which is critical to a specific class, we employ proposals' confidence to each class as the weight during merging, and prototypes are derived as the weighted mean embedding of region proposals:
\begin{equation} \label{eq5}
c_k = \frac{\sum_{i=1}^{N_p} \, \widetilde{\textbf{P}}_{ik} \cdot \widetilde{\textbf{F}}^T_i}{\sum_{i=1}^{N_p} \, \widetilde{\textbf{P}}_{ik}} ,
\end{equation}
where $c_k \in \mathbb{R}^d$ denotes the prototype of class $k$. The derived prototypes serve as the proxy of each class during subsequent domain alignment. 


\par{\textbf{Category-level domain alignment.}} Prototype-based domain alignment is comprehensively studied in recent literatures \cite{semantic,transferrable_proto,minimax_entropy}. The core idea of these methods is to narrow the distance between same categories' prototypes of two domains, which is achieved through minimizing an intra-class loss, noted as $\mathcal{L}_{intra}$. Furthermore, we propose that the distance between different classes' prototypes should also be constrained with another inter-class loss, noted as $\mathcal{L}_{inter}$. In addition, considering the existence of class-imbalance, the influence of different classes needs to be adjusted. The detailed training scheme is presented in the next section.


\subsection{Class-imbalance-aware Adaptation Training} \label{sec3_3}

In object detection scenario, the class-imbalance problem commonly exists, which means the number of samples belonging to different classes varies greatly. Former work \cite{focal_loss} deems that such problem can overwhelm training and degrade detector's performance. In cross-domain detection tasks, class-imbalance can lead to another trouble: the domain adaptation process among different classes is highly unbalanced. In particular, the feature distributions of sample-scarce categories can't be readily aligned. Inspired by Focal Loss \cite{focal_loss} which puts more weights on hard-to-classify examples, we would like to assign higher weights to the sample-scarce categories during the training process of domain adaptation.

Considering that the categories with abundant samples are trained more sufficiently and better aligned, especially in the early training phase, they should possess higher confidence compared with sample-scarce categories. 
Based on this fact, we select a specific class's highest confidence in a set of proposals, and such confidence value is employed to calculate the weight of this class:
\begin{equation} \label{eq6}
p_k = \max \limits_{1 \leqslant i \leqslant N_p} \{ \widetilde{\textbf{P}}_{ik} \} ,
\end{equation}
\begin{equation} \label{eq7}
\alpha_{k} = \left\{
\begin{array}{ll}
(1 - p_k)^{\gamma} & \textrm{if} \ \ p_k > \frac{1}{N_c} \\
0 & \textrm{otherwise} \\
\end{array} \right. ,
\end{equation}
where $p_k$ is the maximum confidence of class $k$ within $N_p$ proposals, and $\gamma$ is the parameter controlling the weights among different classes. Also, we apply a hard threshold, $1 / N_c$, to filter out those classes whose samples are not included in the proposal set. 

Contrastive loss \cite{contrastive_loss} is commonly used in siamese network architecture to enhance the intra-class compactness and inter-class separability. Utilizing such property, we propose a \emph{Class-reweighted Contrastive Loss} to conduct domain alignment on category level, in which class weights $\{\alpha_i^{\mathcal{S}}\}_{i=0}^{N_c}$ and $\{\alpha_i^{\mathcal{T}}\}_{i=0}^{N_c}$ reweight each term in the loss (``$i=0$'' denotes background). Concretely, in this loss function, the intra-class part requires identical classes' prototypes to be as close as possible, and the inter-class part constrains the distance between different classes' prototypes to be larger than a margin:
\begin{small}
	\begin{equation} \label{eq8}
	\mathcal{L}_{intra} (\mathcal{S}, \mathcal{T}) = \frac{\sum_{i=0}^{N_c} \, \alpha_i^{\mathcal{S}} \alpha_i^{\mathcal{T}} \Phi(c_i^{\mathcal{S}}, c_i^{\mathcal{T}})}{\sum_{i=0}^{N_c} \, \alpha_i^{\mathcal{S}} \alpha_i^{\mathcal{T}}} ,
	\end{equation}
	\begin{equation} \label{eq9}
	\mathcal{L}_{inter} (\mathcal{D}, \mathcal{D}') = \frac{\sum \limits_{0 \leqslant i \neq j \leqslant N_c} \! \! \alpha_i^{\mathcal{D}} \alpha_j^{\mathcal{D}'} max(0, m - \Phi(c_i^{\mathcal{D}}, c_j^{\mathcal{D}'}))}{\sum \limits_{0 \leqslant i \neq j \leqslant N_c} \! \! \alpha_i^{\mathcal{D}} \alpha_j^{\mathcal{D}'}} ,
	\end{equation}
	\begin{equation} \label{eq10}
	\begin{split}
	\mathcal{L}_{da} = \ \mathcal{L}_{intra} & (\mathcal{S}, \mathcal{T}) + \frac{1}{3} \, \big( \mathcal{L}_{inter} (\mathcal{S}, \mathcal{S}) \\
	& + \mathcal{L}_{inter} (\mathcal{S}, \mathcal{T}) + \mathcal{L}_{inter} (\mathcal{T}, \mathcal{T}) \big) , 
	\end{split}
	\end{equation}
\end{small}
where $\Phi(x,x') = ||x - x'||_2$ calculates the Euclidean distance between two prototypes, and $\{c_i^{\mathcal{S}}\}_{i=0}^{N_c}$, $\{c_i^{\mathcal{T}}\}_{i=0}^{N_c}$ denote the prototypes of source and target domain. $\mathcal{D}$ and $\mathcal{D}'$ represent two domains from which pairs of prototypes belonging to different categories are taken. $m$ is the margin term which is fixed as $1.0$ in all experiments. In the total domain adaptation loss $\mathcal{L}_{da}$,
all pairwise relations between two domains' prototypes are considered.


\begin{table*}[t]
	\begin{spacing}{1.1}
		\centering
		\small
		\caption{Experimental results (\%) of \emph{Normal to Foggy} cross-domain detection task, Cityscapes $\rightarrow$ Foggy Cityscapes.} \label{fog_city}
		\setlength{\tabcolsep}{3.2mm}
		\begin{tabular}{c|cccccccc|c}
			\toprule[1.0pt]
			Methods & person & rider & car & truck & bus & train & motorcycle & bicycle & mAP \\
			\hline
			\hline
			Source-only & 26.9 & 38.2 & 35.6 & 18.3 & 32.4 & 9.6 & 25.8 & 28.6 & 26.9 \\
			DA \cite{da_faster_rcnn} & 29.2 & 40.4 & 43.4 & 19.7 & 38.3 & 28.5 & 23.7 & 32.7 & 32.0 \\
			DivMatch \cite{diversify_and_match} & 31.8 & 40.5 & 51.0 & 20.9 & 41.8 & 34.3 & 26.6 & 32.4 & 34.9 \\
			SW-DA \cite{strong-weak} & 31.8 & 44.3 & 48.9 & 21.0 & 43.8 & 28.0 & 28.9 & 35.8 & 35.3 \\
			SC-DA \cite{adapting_detector} & \textbf{33.8} & 42.1 & 52.1 & \textbf{26.8} & 42.5 & 26.5 & 29.2 & 34.5 & 35.9 \\
			MTOR \cite{mean_teacher} & 30.6 & 41.4 & 44.0 & 21.9 & 38.6 & 40.6 & 28.3 & 35.6 & 35.1 \\
			\hline
			\hline
			GPA (RPN Alignment)  & 32.5 & 43.1 & 53.3 & 22.7 & 41.4 & 40.8 & 29.4 & 36.4 & 37.4  \\
			GPA (RCNN Alignment) & 33.5 & 44.8 & 52.6 & 26.0 & 41.2 & 37.6 & 29.8 & 35.2 & 37.6 \\
			GPA (Two-stage Alignment) & 32.9 & \textbf{46.7} & \textbf{54.1} & 24.7 & \textbf{45.7} & \textbf{41.1} & \textbf{32.4} & \textbf{38.7} & \textbf{39.5} \\
			\bottomrule[1.0pt]
		\end{tabular}
	\end{spacing}
	\vspace{-2mm}
\end{table*}


\subsection{Two-stage Domain Alignment} \label{sec3_4}

Faster R-CNN \cite{faster_rcnn} is a two-stage object detector made up of Region Proposal Network (RPN) and Region-based CNN (R-CNN). 
First, based on the feature map produced by bottom convolutional layers, RPN generates class-agnostic region proposals. After that, R-CNN predicts fine-grained category labels from feature vectors obtained via ROI pooling. Each stage defines a classification and a localization error, and the total detection loss is defined as follows:

\vspace{-0.3cm}
\begin{equation} \label{eq11}
\mathcal{L}_{det} = \mathcal{L}^{RPN}_{cls} + \mathcal{L}^{RPN}_{loc} + \mathcal{L}^{RCNN}_{cls} + \mathcal{L}^{RCNN}_{loc} .
\end{equation}

Based on the two-stage structure of Faster R-CNN, we also conduct domain alignment in a two-stage manner. In the first stage, using the region proposals and corresponding class-agnostic confidence produced by RPN, foreground and background features are separated on latent space, and the foreground feature distributions are aligned as a whole. In the second stage, by utilizing the more accurate bounding boxes and per-category confidence, the feature distribution of each category is respectively aligned. Applying the proposed Class-reweighted Contrastive Loss to both RPN and RCNN, the overall objective is:
\begin{equation} \label{eq12}
\min \limits_{F_{\theta}} \ \mathcal{L}_{det} + \lambda_1 \mathcal{L}^{RPN}_{da} + \lambda_2  \mathcal{L}^{RCNN}_{da} , 
\end{equation}
where $F_{\theta}$ represents the whole parameterized model, and $\lambda_1$ and $\lambda_2$ are the trade-off parameters between detection and domain adaptation loss.

\par{\textbf{Implementation details.}} On the basis of ResNet-50 \cite{resnet} architecture, we implement two domain adaptation losses, $\mathcal{L}^{RPN}_{da}$ and $\mathcal{L}^{RCNN}_{da}$, through adding two domain adaptation learning branches to the $7 \times 7 \times 1024$ feature map after ROI pooling and the $2048$-dimensional vector after average pooling, respectively. 


\section{Experiments} \label{sec4}

In this section, we provide comprehensive experimental results on three cross-domain detection tasks with distinct domain shift, including \emph{Normal to Foggy}, \emph{Synthetic to Real} and \emph{Cross Camera Adaptation}. 


\subsection{Experimental Setup} \label{sec4_1}

\par{\textbf{Training details.}} In all experiments, unless otherwise specified, all of the training and test images are resized such that their shorter side has 600 pixels. During training, for each image, 128 anchors are sampled with a positive to negative ratio of $1:3$. 
ResNet-50 \cite{resnet} pre-trained on ImageNet \cite{imagenet} serves as the base architecture. We adopt the SGD optimizer (initial learning rate: 0.001, momentum: 0.9, weight decay: $5 \times 10^{-4}$) to train our model. The number of total training epoch is set as 20, and the learning rate warm-up strategy \cite{resnet} is used in the first 200 iterations of training. Without specific notation, the class-balancing hyper-parameter $\gamma$ is set as 2.0, and the IoU-based adjacency matrix defined in Eq. \ref{eq2} is adopted. For evaluation, we report mean average precisions (mAP) with a threshold of 0.5.

In our experiments, two NVIDIA GeForce 1080 Ti GPUs are used for training, and we select the batch size of 12 to fit GPU memory, \emph{i.e.} 6 images per GPU, consisting of 3 labeled samples from source domain and 3 unlabeled samples from target domain. Our method is implemented with the PyTorch \cite{pytorch} deep learning framework. 


\par{\textbf{Performance comparison.}} We compare our approach with state-of-the-art methods to verify its effectiveness. Our method is evaluated under three configurations corresponding to RPN Alignment ($\lambda_1 = 1.0$, $\lambda_2 = 0.0$), RCNN Alignment ($\lambda_1 = 0.0$, $\lambda_2 = 1.0$) and Two-stage Alignment ($\lambda_1 = 1.0$, $\lambda_2 = 1.0$). Former works, DA \cite{da_faster_rcnn}, DivMatch \cite{diversify_and_match}, SW-DA \cite{strong-weak}, SC-DA \cite{adapting_detector} and MTOR \cite{mean_teacher} are introduced for comparison. For the sake of fair comparison, we employ ResNet-50 as the backbone for all these methods. In specific, we re-evaluate the performance of DA, DivMatch, SW-DA and SC-DA using their source code with default configuration, and the performance of MTOR in original paper is reported for the lack of source code.


\subsection{Normal to Foggy} \label{sec4_2}

\par{\textbf{Datasets.}} In this experiment, Cityscapes \cite{city} and Foggy Cityscapes \cite{foggy} dataset serve as source and target domain, respectively. Cityscapes dataset contains 2,975 training images and 500 validation images, and we follow the operation in \cite{da_faster_rcnn} to get the detection annotations. Foggy Cityscapes dataset simulates fog on real scenes through rendering the images from Cityscapes, and it shares the same annotations with Cityscapes dataset. 
The results are reported on the validation set of Foggy Cityscapes. 

\par{\textbf{Results.}} In Table \ref{fog_city}, the comparisons between our approach and other cross-domain detection methods are presented on eight categories. Source-only denotes the baseline Faster R-CNN trained with only source domain data. From the table, it can be observed that the performance of our approach under three configurations all surpasses existing methods. In particular, an increase of $3.6\%$ on mAP is achieved by Two-stage Alignment. The results showcase that, under the domain shift caused by local fog noise, the proposed \emph{graph-based region aggregation} can effectively alleviate such noise and extract critical instance-level features. Take a closer look at per-category performance, our approach achieves highest AP on most sample-scarce categories, \emph{i.e.} rider, bus, train and motorcycle. This phenomenon illustrates the effectiveness of \emph{Class-imbalance-aware Adaptation Training} on balancing the domain adaptation process among different classes. 


\subsection{Synthetic to Real} \label{sec4_3}

\par{\textbf{Datasets.}} In this experiment, SIM 10k \cite{sim10k} dataset is employed as source domain. SIM 10k dataset is collected from the computer game Grand Theft Auto V (GTA5), and it contains 10,000 images. Cityscapes \cite{city} dataset serves as target domain, and experimental results are reported on its validation split. 

\par{\textbf{Results.}} Table \ref{sim10k_city} reports the performance of our approach compared with other works on two datasets' common category, car. The Two-stage Alignment configuration of our approach obtains the highest AP ($47.6\%$) over all methods. The domain shift of this task is mainly brought by distinct image styles. In such case, in order to achieve satisfactory performance, it's important to produce discriminative features between foreground and background on target domain. We think that, in our framework, such goal is realized through constraining inter-class separability in the \emph{Class-reweighted Contrastive Loss}.


\begin{table}[t]
	\begin{spacing}{1.1}
		\centering
		\small
		\caption{Experimental results (\%) of \emph{Synthetic to Real} cross-domain detection task, SIM 10k $\rightarrow$ Cityscapes.} \label{sim10k_city}
		\setlength{\tabcolsep}{8mm}
		\begin{tabular}{c|c} 
			\toprule[1.0pt]
			Methods & \emph{car} AP \\
			\hline
			\hline
			Source-only & 34.6 \\
			DA \cite{da_faster_rcnn} & 41.9 \\
			DivMatch \cite{diversify_and_match} & 43.9 \\
			SW-DA \cite{strong-weak} & 44.6 \\
			SC-DA \cite{adapting_detector} & 45.1 \\
			MTOR \cite{mean_teacher} & 46.6 \\
			\hline
			\hline
			GPA (RPN Alignment)  & 45.1\\
			GPA (RCNN Alignment)  & 44.8 \\
			GPA (Two-stage Alignment)  & \textbf{47.6} \\
			\bottomrule[1.0pt]
		\end{tabular}
	\end{spacing}
	\vspace{-1mm}
\end{table}


\subsection{Cross Camera Adaptation} \label{sec4_4}

\textbf{Datasets.} In this part, we want to explore the adaptation between real-world datasets under different camera setups. KITTI \cite{kitti} dataset serves as source domain, and it contains 7,481 training images.
Cityscapes \cite{city} dataset is utilized as target domain, and its validation set is used for evaluation.


\textbf{Results.} The results of various methods on two datasets' common category, car, are presented in Table \ref{kitti_city}. In this task, all three configurations of our approach exceed existing works with a notable margin, in particular, $4.3\%$ performance gain achieved by Two-stage Alignment. In cross camera adaptation tasks, due to the difference of camera setups, abundant patterns exist in instances. In our method, the multi-modal information reflected by various instances is integrated into prototype representations, such that the diverse patterns within a specific category are considered during domain adaptation, which promises the superior performance of our approach.


\section{Analysis}  \label{sec5}

In this section, we provide more in-depth analysis of our approach to validate the effectiveness of major components with both quantitative and qualitative results.


\subsection{Ablation Study} \label{sec5_1}


\par{\textbf{Effect of relation graph.}} In Table \ref{ablation_graph}, we analyze a key component, \emph{i.e.} the relation graph, on the task SIM 10k $\rightarrow$ Cityscapes. The first row directly uses the original region proposals produced by RPN to compute prototypes, and it serves as the baseline. In the second row, we use an Euclidean distance based relation graph defined in Eq. \ref{eq1}, in which $\sigma$ is set as $15.0$ so as to keep the sparsity of derived relation graph same as the one defined by IoU. Comparing the second and fourth row, it can be observed that the configuration using IoU based relation graph performs better, which illustrates that region proposals' size information is essential for relation graph construction.


\begin{table}[t]
	\begin{spacing}{1.1}
		\centering
		\small
		\caption{Experimental results (\%) of \emph{Cross Camera Adaptation} task, KITTI $\rightarrow$ Cityscapes.} \label{kitti_city}
		\setlength{\tabcolsep}{8mm}
		\begin{tabular}{c|c}
			\toprule[1.0pt]
			Methods & \emph{car} AP \\
			\hline
			\hline
			Source-only & 37.6 \\
			DA \cite{da_faster_rcnn} & 41.8 \\
			DivMatch \cite{diversify_and_match} & 42.7 \\
			SW-DA \cite{strong-weak} & 43.2 \\
			SC-DA \cite{adapting_detector} & 43.6 \\
			\hline
			\hline
			GPA (RPN Alignment)  & 46.9 \\
			GPA (RCNN Alignment)  & 46.1 \\
			GPA (Two-stage Alignment)  & \textbf{47.9} \\
			\bottomrule[1.0pt]
		\end{tabular}
	\end{spacing}
\end{table}


\begin{table}[t]
	\begin{spacing}{1.1}
		\centering
		\small
		\caption{Ablation study on different manners to construct relation graph. (``ED'': Euclidean distance, ``LP'': learnable parameter.)}
		\label{ablation_graph}
		\setlength{\tabcolsep}{7mm}
		\begin{tabular}{ccc|c}
			\toprule[1.0pt]
			ED & IoU & LP & \emph{car} AP \\
			\hline
			&  &  & 45.0 \\
			\checkmark &  &  & 46.1 \\
			\checkmark &  & \checkmark & 43.2 \\
			& \checkmark &  & 47.6 \\
			& \checkmark & \checkmark & 43.6 \\
			\bottomrule[1.0pt]
		\end{tabular}
	\end{spacing}
	\vspace{-1mm}
\end{table}


\begin{figure*}[t]
	\centering
	\includegraphics[width=.97\textwidth]{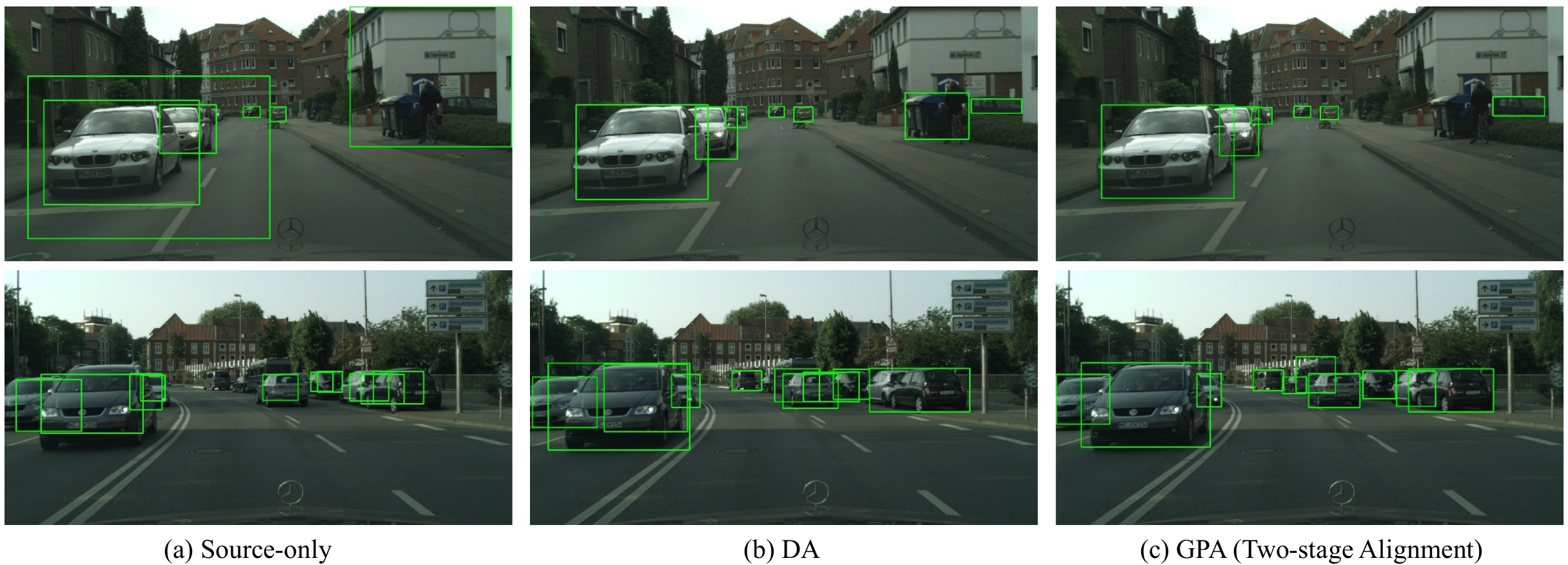}
	\caption{The detection results on the task SIM 10k $\rightarrow$ Cityscapes, in which Source-only, DA \cite{da_faster_rcnn} and our method are evaluated.} 
	\label{fig_detection_results}
	\vspace{-3mm}
\end{figure*}


\begin{figure}[t]
	\centering
	\includegraphics[width=0.48\textwidth]{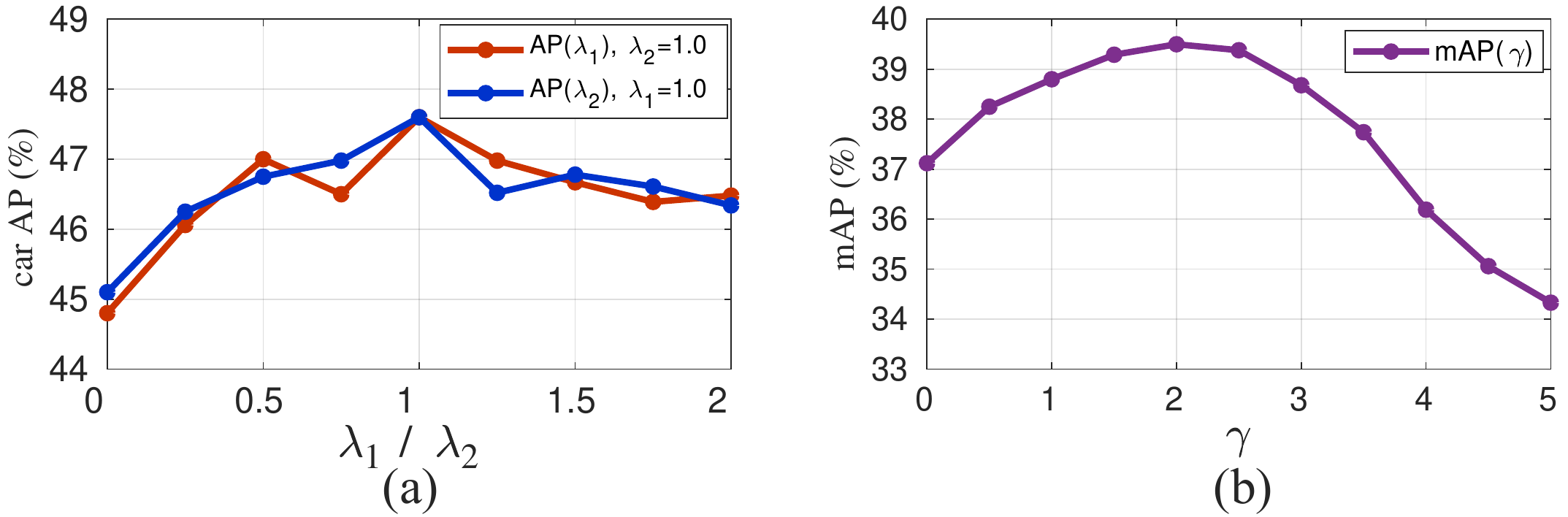}
	\caption{Sensitivity analysis of trade-off parameters $\lambda_1$, $\lambda_2$ (left) and class-balancing parameter $\gamma$ (right).} 
	\label{sensitivity_analysis}
	\vspace{-2mm}
\end{figure}


In the third and fifth row, we append the learnable parameter matrix to Eqs. \ref{eq3}, \ref{eq4}, which forms the conventional formula of graph convolution. After introducing such learnable parameter matrix, compared with the parameter-free counterparts in the second and fourth row, apparent performance decay occurs. We suppose that such phenomenon can be ascribed to the lack of explicit supervisory signal on the branch of domain adaptation learning, which makes it hard to learn a proper feature transformation.


\par{\textbf{Effect of two-stage alignment.}} In this part, we demonstrate the effectiveness of two-stage alignment. In different cross-domain detection tasks, as shown in Table \ref{fog_city}, \ref{sim10k_city} and \ref{kitti_city}, three configurations of the proposed approach are evaluated. Two single-stage configurations possess similar performance, and two-stage alignment surpasses them with a clear margin. These results illustrate that two-stage alignment boosts domain adaptation via a progressive alignment manner, \emph{i.e.} from coarse-grained foreground alignment to fine-grained per-category alignment. 


\subsection{Sensitivity Analysis} \label{sec5_2}


\par{\textbf{Sensitivity of trade-off parameters $\lambda_1$, $\lambda_2$.}} In this experiment, we validate our approach's sensitivity to $\lambda_1$ and $\lambda_2$ which trade off between detection and domain adaptation loss. Figure \ref{sensitivity_analysis}(a) shows model's performance under different $\lambda_1$ ($\lambda_2$) values when the other parameter $\lambda_2$ ($\lambda_1$) is fixed, and all results are evaluated on the task SIM 10k $\rightarrow$ Cityscapes. From the line chart, it can be observed that the performance on target domain is not sensitive to both parameters when they vary from $0.25$ to $2.0$, and apparent performance gain is obtained compared with RCNN Alignment ($\lambda_1 = 0$) and RPN Alignment ($\lambda_2 = 0$). This phenomenon illustrates that the two-stage alignment can achieve satisfactory results on a wide range of trade-off parameters. 


\par{\textbf{Sensitivity of class-balancing parameter $\gamma$.}} In this part, we discuss the selection of parameter $\gamma$ which balances the domain adaptation process among different categories. In Figure \ref{sensitivity_analysis}(b), we plot the performance of models trained with different $\gamma$ value on the task Cityscapes $\rightarrow$ Foggy Cityscapes. The highest mAP on target domain is achieved when the value of $\gamma$ is around $2.0$, which means that, under such condition, the weight assignment among different classes benefits domain adaptation most. 


\subsection{Visualization} \label{sec5_3}


\par
\begin{figure}[t]
	\centering
	\includegraphics[width=0.47\textwidth]{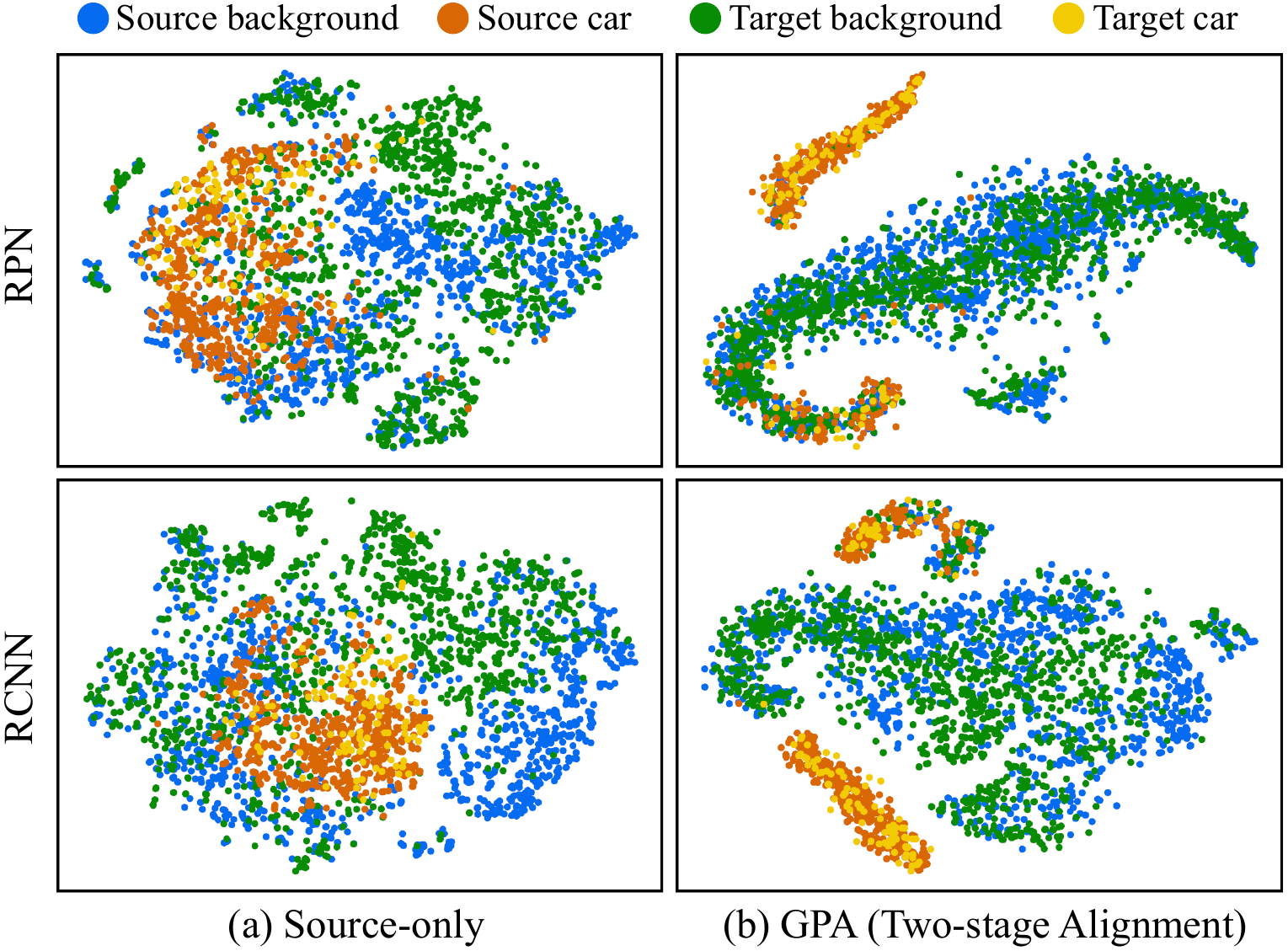}
	\caption{The t-SNE \cite{tsne} visualization of feature embeddings produced by RPN and RCNN, in which Source-only model and our method are employed for feature extraction.} 
	\label{fig_tsne}
	\vspace{-3.5mm}
\end{figure}

{\textbf{Visualization of two-stage feature.}} In Figure \ref{fig_tsne}, we utilize t-SNE \cite{tsne} to visualize the feature distribution of source and target domain on the task SIM 10k $\rightarrow$ Cityscapes, in which the feature embeddings of both RPN and RCNN phase are used for visualization. Compared with the Source-only model, after conducting RPN and RCNN alignment, the features of the same category in two domains are better aligned, and different categories' features are separated more clearly. This visually verifies that the proposed method boosts feature alignment on both stages. 


\par{\textbf{Qualitative detection results.}} Figure \ref{fig_detection_results} displays some typical detection results on the task SIM 10k $\rightarrow$ Cityscapes, in which Source-only, DA \cite{da_faster_rcnn} and our approach are evaluated. As shown in the figure, the Source-only model can poorly localize objects. DA \cite{da_faster_rcnn} predicts bounding box more accurately, but it incorrectly classifies the garbage can as a car, and produces some false positives. Our model successfully inhibits false positives, and it is able to localize objects precisely even when severe occlusion occurs. 


\section{Conclusion}  \label{sec6}

In this paper, we propose the Graph-induced Prototype Alignment (GPA) framework for cross-domain detection. In the framework, the critical information of each instance is aggregated through graph-based message propagation, and prototype representations are derived for category-level domain alignment. Furthermore, we harmonize the process of adaptation training through Class-reweighted Contrastive Loss. Extensive experiments and analytical studies demonstrate the prominent performance of our approach.


\vspace{-0.5mm}
\section{Acknowledgement} \label{sec7}

This work was supported by National Science Foundation of China (61976137, U1611461, U19B2035) and STCSM(18DZ1112300). This work was also supported by National Key Research and Development Program of China (2016YFB1001003). Authors would like to appreciate the Student Innovation Center of SJTU for providing GPUs.


\newpage
{\small
\bibliographystyle{ieee_fullname}
\bibliography{reference}
}
\newpage


\begin{figure}[t]
	\centering
	\includegraphics[width=0.99\columnwidth]{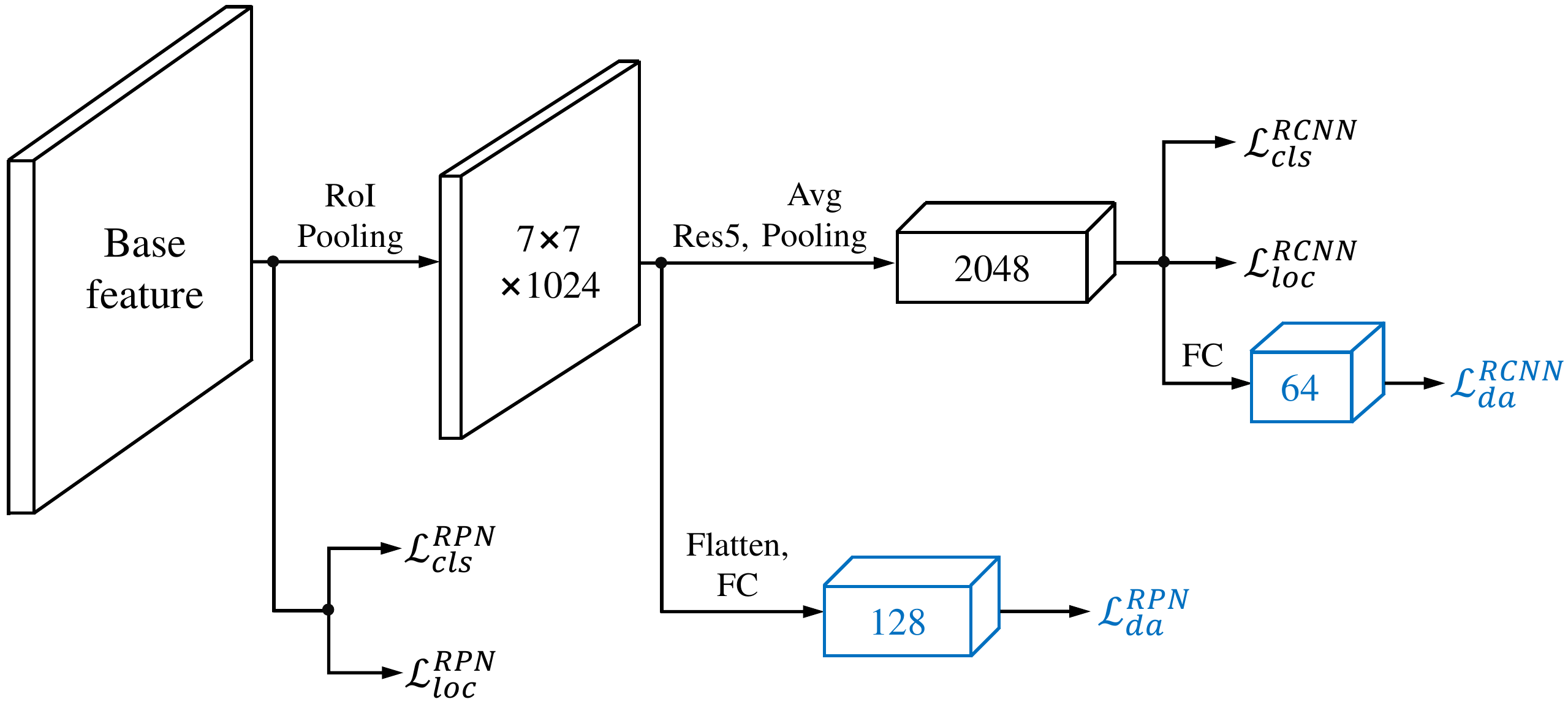}
	\caption{The architecture of \emph{head} network. It is built on the basis of Faster R-CNN \cite{faster_rcnn} with ResNet-50 \cite{resnet} backbone.} \label{fig_network}
\end{figure}


\section{Appendix \uppercase\expandafter{\romannumeral1}: Network Architecture} \label{appendix1}

In this work, we instantiate the proposed Graph-induced Prototype Alignment (GPA) framework using Faster R-CNN \cite{faster_rcnn} detector with ResNet-50 \cite{resnet} backbone. For clarity, we split the whole network architecture into two parts: (1) the \emph{backbone} network for feature extraction over entire images, and (2) the \emph{head} network for bounding box recognition (classification and regression) and domain adaptation learning, which is presented in Figure \ref{fig_network}.

The whole framework is composed of two stages, Region Proposal Network (RPN) and Region-based CNN (RCNN). 
For RPN, by utilizing the base feature extracted with ResNet-50 backbone, the classification and localization losses, $\mathcal{L}^{RPN}_{cls}$ and $\mathcal{L}^{RPN}_{loc}$, are defined, and the $7 \times 7 \times 1024$ feature map of each region proposal is generated through RoI pooling. After flattening the feature map, a fully-connected layer outputs the 128-dimensional feature vector which derives foreground and background prototypes, and the domain alignment loss $\mathcal{L}^{RPN}_{da}$ is calculated with these prototypes. For RCNN, a 2048-dimensional feature vector is generated via average pooling, and the classification and localization losses, $\mathcal{L}^{RCNN}_{cls}$ and $\mathcal{L}^{RCNN}_{loc}$, are defined on such basis. By using another fully-connected layer, the 64-dimensional feature vector is produced to derive prototypes of each category, and, based on these prototypes, the domain alignment loss $\mathcal{L}^{RCNN}_{da}$ is calculated. 


\section{Appendix \uppercase\expandafter{\romannumeral2}: Qualitative Detection Results} \label{appendix2}

\begin{figure*}[t]
	\centering
	\includegraphics[width=.92\textwidth]{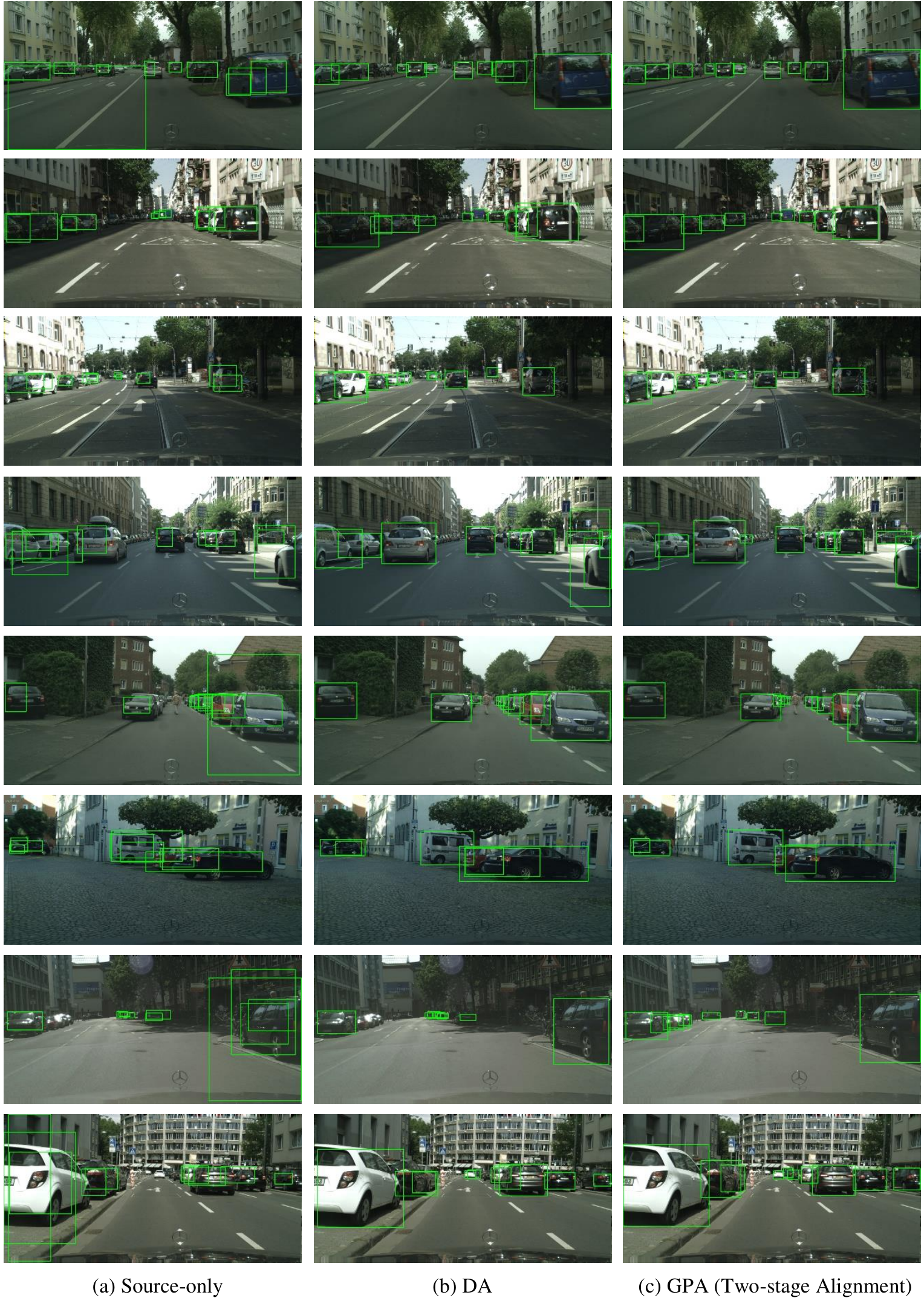}
	\caption{The detection results on the task SIM 10k $\rightarrow$ Cityscapes, in which Source-only, DA \cite{da_faster_rcnn} and our method are evaluated.} 
	\label{car_more_results}
	\vspace{-3mm}
\end{figure*}


\begin{figure*}[t]
	\centering
	\includegraphics[width=.90\textwidth]{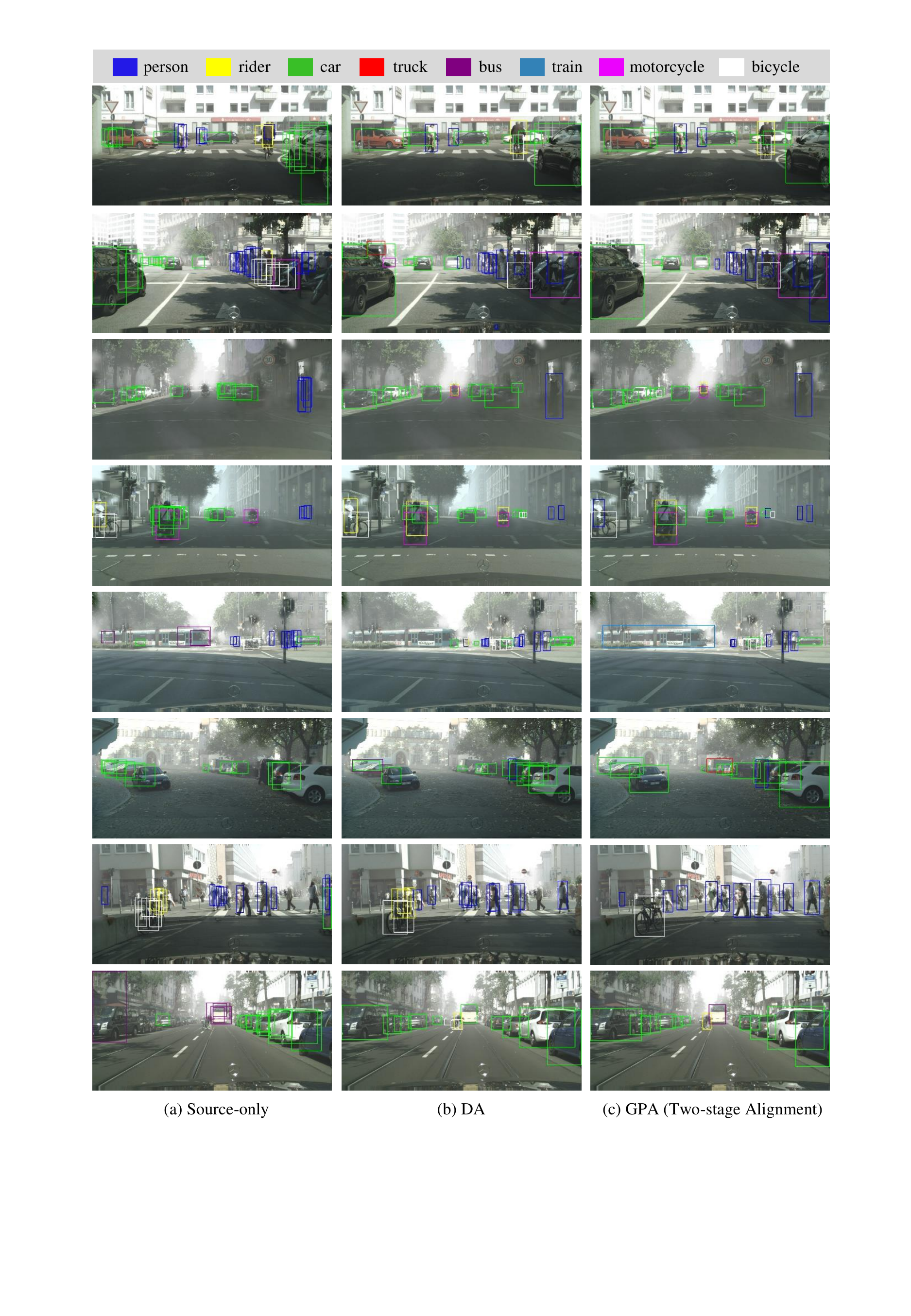}
	\caption{The detection results on the task Cityscapes $\rightarrow$ Foggy Cityscapes, in which Source-only, DA \cite{da_faster_rcnn} and our method are evaluated.} 
	\label{multi_class_more_results}
	\vspace{-3mm}
\end{figure*}


In Figure \ref{car_more_results}, we present more detection results on the task SIM 10k $\rightarrow$ Cityscapes, and this task aims for vehicle detection. As shown in the figure, the Source-only model produces many bounding boxes greatly biasing from objects, since the generated features are not discriminative enough. DA \cite{da_faster_rcnn} localizes objects more precisely, but some false positives are produced by this method, \emph{e.g.} the second figure of the fifth and sixth rows. In the results of our approach, these false positives are effectively alleviated, and our model can accurately localize those objects with small scale and in severe occlusion, \emph{e.g.} the third figure of the first row.

Figure \ref{multi_class_more_results} displays several groups of detection results on the task Cityscapes $\rightarrow$ Foggy Cityscapes. On this task, eight common categories of two datasets are used for evaluation. In the results of Source-only and DA \cite{da_faster_rcnn}, quite a few bounding boxes are assigned with false labels, and several objects are undetected. For example, in the last row, a bus is misclassified as car by the DA model. Our approach correctly detects most of the objects and predicts bounding boxes more accurately. Just as shown in the fifth row, a train is undetected using Source-only and DA model, while it is precisely localized by our method.  


\end{document}